# Prototyping Virtual Reality Serious Games for Building Earthquake Preparedness: The Auckland City Hospital Case Study

Ruggiero Lovreglio, Vicente Gonzalez, Zhenan Feng, Robert Amor, Michael Spearpoint, Jared Thomas, Margaret Trotter, Rafael Sacks

**Abstract:** Enhancing evacuee safety is a key factor in reducing the number of injuries and deaths that result from earthquakes. One way this can be achieved is by training occupants. Virtual Reality (VR) and Serious Games (SGs), represent novel techniques that may overcome the limitations of traditional training approaches. VR and SGs have been examined in the fire emergency context; however, their application to earthquake preparedness has not yet been extensively examined.
We provide a theoretical discussion of the advantages and limitations of using VR SGs to investigate how building occupants behave during earthquake evacuations and to train building occupants to cope with such emergencies. We explore key design components for developing a VR SG framework: (a) what features constitute an earthquake event; (b) which building types can be selected and represented within the VR environment; (c) how damage to the building can be determined and represented; (d) how non-player characters (NPC) can be designed; and (e) what level of interaction there can be between NPC and the human participants. We illustrate the above by presenting the Auckland City Hospital, New Zealand as a case study, and propose a possible VR SG training tool to enhance earthquake preparedness in public buildings.

**Abbreviations:**
VR: Virtual Reality
SG: Serious Game
BP: Behavioural Prototype
TP: Training Prototype
ACH: Auckland City Hospital
HMD: Head Mounted Display
NPC: Non-player Character

# 1. INTRODUCTION

Dropping, covering and holding during an earthquake followed by careful building evacuation once it is safe to do so is the main response strategy recommended in several earthquake-prone countries, such as New Zealand, the United States, Japan and Italy [1–3]. Population awareness of these recommended behaviours is paramount to increasing building occupants' chance of survival.

Two risk-reduction approaches have been proposed to enhance evacuee safety and thus to reduce the number of injuries and deaths due to earthquakes. The first uses a 'behavioural design' approach which involves risk-reduction interventions in new and existing buildings [4]. This strategy relies on the existing knowledge of how building occupants would behave in earthquake evacuations. The second enhances building occupants' earthquake evacuation preparedness through training. Building occupants are instructed on how to get ready for an earthquake, and how to behave during and after an earthquake, by following the recommended behaviours [5]. Ultimately, these two risk-reduction approaches highlight the importance of being able to predict human behaviour in earthquakes and to train building occupants.

Different approaches have been proposed to investigate evacuation behaviour and train building occupants [4]. Evacuation drills are the most used traditional approach especially in fire emergencies. The main limitation of evacuation drills is the difference between the real-world emergency and the simulated emergency, which can significantly prevent participants' learning [6,7]. It is clear that there is a difference between fire and earthquake conditions. In many real fire evacuations, it is quite possible that most evacuees do not even see fire and smoke. Therefore, the gap of realism between real fire evacuations and 'classic' fire drills can be almost insignificant. In contrast, in earthquake evacuations everyone can perceive the threat. Therefore, there is a significant gap between real evacuations and earthquake drills that could be reduced using Virtual Reality (VR) and Serious Games (SGs). Moreover, evacuation drills might trigger evacuation behaviours that are different from those observed in real earthquake evacuations [8]. From a pedagogical point of view, it is difficult to ensure that evacuation drills provide effective training. In fact, evacuation drill participants often receive no feedback whatsoever to help them assess their evacuation choices retrospectively [6,7].

VR and SGs represent novel and effective techniques to overcome the limitations of traditional approaches [9]. VR technologies (i.e. tools to immerse users in a computer generated synthetic environment) have proven to be valuable alternatives to investigate several behaviours in fire evacuation, such as system perception [10,11], pre-movement behaviour, wayfinding, exit choice [12,13] and navigation interactions [14,15]. This emerging technology allows users to be exposed to more realistic evacuation scenarios by allowing the representation of several threats; in the earthquake context, this includes building damage [16]. SG concepts have been used over recent decades to train people to cope with different fire emergency issues [17] and several evacuation scenarios [18–20]. Nevertheless, despite the advantages provided by VR and SGs, applications in the space of earthquake evacuations are still rare. There are only a few VR applications related to earthquake preparedness [21], as described in section 2.

The objective of this paper is to discuss the advantages and limitations of the combined use of VR and SGs to understand building occupant behaviour in the event of an earthquake and to develop a training tool to teach building occupants to cope effectively (during and after). The paper presents a case study showing the practical steps required to develop a VR SG prototype to improve post-earthquake evacuation preparedness. Thus, we discuss the pipelines used to develop two VR SG prototypes. We will refer to the first, aimed at investigating human behaviour during an earthquake event, as the 'Behavioural Prototype' (BP), while we will call the second, aimed at teaching users the best earthquake evacuation practice according to New Zealand Civil Defence [2], the 'Training Prototype' (TP).

## 2. VIRTUAL REALITY AND SERIOUS GAMES

SGs represent a new way to investigate human behaviour and to train building occupants to cope with earthquake evacuations. To date, several definitions have been provided and discussed for SGs. For instance, Michal and Chen [22] (p. 17) define a SG as *"a game in which education (in its various forms) is the primary goal, rather than entertainment",* highlighting that education and entertainment are not in conflict but overlap, enhancing the outcomes of the experience for the users.

SGs can be developed using any existing game engine [23]. To date, there are more than 600 commercial game engines and the choice amongst all of them depends on the type of SG (i.e. the world of play) as well as the features that need to be included in the game. In many cases, several game engines can be identified to achieve the identical gaming goal [23]. Therefore, the choice of one game engine over others may depend on other secondary constraints (i.e. costs, supported libraries and languages, etc.).

SGs have been tested for training purposes in fire emergencies [17] and for enhancing users' earthquake preparedness [21]. For instance, Tanes and Cho [24] investigated the possibility of preparing people to take precautions against earthquakes by using the online game entitled *'Beat the Quake!'* developed by the Earthquake Country Alliance. Takimoto et al. [25] and Dunwell et al. [26] aimed to provide users with knowledge of safe evacuation processes and procedures within game environments. The main limitation of those applications is the low level of realism and related level of engagement.

There are only two applications combining VR and SG technologies in the literature. Li et al. [27] developed several common indoor earthquake scenarios to teach individuals how to survive earthquakes. This study indicates that such a VR training approach is effective. However, the authors only train users in the drop, cover and hold actions, without providing any recommended actions to safely evacuate a building after an earthquake. As such, the authors address VR SG development issues related to a single building rather than a section of a building. The second VR application is "Earthquake Simulator VR" and was developed by Lindero Edutainment for HTC VIVE headsets [28]. This VR application has several training objectives as it teaches users (1) how to prepare an emergency preparedness kit before an earthquake; (2) how to drop, cover and hold during an earthquake; and (3) how to survive a post-earthquake fire by evacuating the building. This application was developed for a simple indoor scenario as the virtual environment consists of a few rooms in a house. This application does not provide users with adequate feedback regarding their actions taken during the experience, so learning is limited. Moreover, there is no validation study showing the training capabilities of this SG. Technical information regarding the development of this application are also not discussed and validated.

Despite the increasing number of SG applications for safety training, the advantages and disadvantages of using such new tools to train building occupants is still under investigation. The main advantage is that SGs can dramatically improve the realism of the learning experience. Features such as debris and unexpectedly blocked exit pathways can easily be included in virtual environments without exposing users to any risk [23,29]. VR simulations have similarly been used for construction safety training, where users can experience hazardous scenarios without risk [30]. Another important advantage is the high control of the evacuation scenarios. Different evacuation scenarios can be designed and used to enhance the training outcomes of the game [18]. However, there is still the need for more longitudinal experimental studies to assess the long-term effects of training using serious games [31]. A potential disadvantage identified by Williams-Bell et al. [17] is that the available gaming technologies are not always capable of accurately representing a virtual environment that mimics the hazard dynamics.

Besides the training purpose of the SGs, these tools also enable investigation of users' behaviours [18]. SGs allow the assessment of users' reactions to different stimuli and their decision-making. For instance, simulating the earthquake damage in the virtual environment can allow the assessment of users' reactions to different evacuation conditions (e.g. different levels of damage and different visibility conditions). Therefore, a SG platform can be seen as a virtual laboratory with higher experimental control than 'traditional' evacuation tools. The validity of such behavioural data can be increased by coupling SGs with new emerging immersive VR technologies [32]. These technologies allow the user to be immersed in a computer-generated virtual environment by using Head Mounted Displays (HMDs) or Cave Automatic Virtual Environments (CAVEs) [33].

The advantage provided by platforms coupling VR and SGs is the amount of behavioural data that can be collected during the game experience. Immersive VR technologies allow investigation of evacuee movement as classic evacuation experiments do. However, one can also monitor evacuee viewing directions and which elements and objects users are looking at before making their decisions [18]. This type of data, which is difficult to collect with classic evacuation experiments such as evacuation drills, can be fundamental to identify what factors affect evacuation behaviour [34–36]. The behavioural outcome can also be increased by using verbal protocol analysis, in which users are encouraged to speak their thoughts aloud while they are engaged in carrying out a task. This psychological research method aims to (1) generate insight into how courses of action are selected and executed and (2) determine the cognitive factors driving people's actions, decision-making and information seeking while dealing with an emergency scenario [37].

Despite the advantages offered by VR technologies for studying human behaviour, there are still technological challenges and limits that need to be addressed [33]. For instance, many HMDs do not allow users to see their own body in the virtual environment, generating a 'ghost experience' [38]. This issue can be solved by combining HMDs with other sensors able to integrate the perception in the virtual environment by using an avatar. Another common issue for HMDs and CAVEs is the limited space to track the movement of users (i.e. trackable areas) [33]. Omnidirectional treadmills, or similar technologies, and game controllers provide a solution that allows navigation in virtual environments to have walkable surfaces bigger than trackable areas. Other open challenges related to immersive VR technologies are motion sickness (such as symptoms of nausea and vertigo), multisensory simulation (i.e. limited tactile and olfactory experiences), and interactivity (i.e. the need for game controllers to interact with objects in the virtual environment) [33].

Motion sickness is indeed the most critical issue to consider while designing a VR experience. This is generated whenever physically stationary VR users see convincing visual representations of self-motion and detectable lags are present between head movements and the presentation of the visual display in helmet-mounted displays [39]. Both situations generate conflicting inputs from visual and vestibular systems, producing motion sickness and postural instability [40]. The lags issues can be addressed by optimizing the VR experience depending on the GUI and CPU capabilities of the hardware system. The sickness related to experiences of self-motion in the absence of actual physical displacement is still a challenge for many VR developers, as predicted in the early 90s [39]. To date, reliable techniques that can fully prevent motion sickness are unfortunately not available. Therefore, suitable VR design solutions must be found on a case-by-case basis [41].

Finally, as argued by Brey [42], the development of VR SGs can raise ethical and moral issues. In virtual earthquake simulations, users are immersed in scenarios that can generate strong emotions (i.e. fear and upset) or recall previous personal dramatic experiences (for instance, whenever a participant has directly experienced an earthquake). This ethical issue points to the need to compromise between the realism of the VR scenario and the emotional reaction that such a VR experience can generate.

## 3. AN EARTHQUAKE GAME ENVIRONMENT

We present two VR SG prototypes for earthquake evacuation: The first, to investigate how building occupants behave, and the second, to train building occupants in coping with such emergencies. The following sections describe the key design questions that were investigated in order to develop these prototypes, namely: (a) what features constitute an earthquake event; (b) how existing and new building should be represented within the VR environment; (c) how damage to the building can be determined and represented; (d) how non-player characters (NPC) can be designed; and (e) what level of interaction there can be between NPC and the human participants.

### 3.1 Earthquake Features

The earthquake event represents the main component that needs to be defined to identify the gaming timeframe for the VR SG prototype. Earthquakes are a sudden and perceptible shaking of the earth's surface and they can generally be characterized by three seismic stages: the foreshock stage, the main shock and the aftershock stage [43]. The sequence of three stages can last several months [44]. This timeframe is not compatible with the duration of a gaming experience. In the context of the VR SG prototypes, however, we identify the earthquake condition as the timeframe that can include at least the main shock and the time required to evacuate the building after the main shock. As such, the gaming timeframe is likely to be of the order of several tens of minutes at most depending on the duration of shocks and the time required to evacuate the building. Considering that the likelihood of foreshock and aftershock events within the timeframe of the main shock is low [45], a trade-off between representativeness of the experience and human factors such as fatigue should be appropriately considered when foreshocks and aftershocks need to be included in the game.

The representation of shakes in a virtual environment denotes a key challenge for the development of a VR SG prototype. An earthquake is characterized by perceptible shakings, while the standard immersive technologies do not allow the inclusion of physical shaking stimuli. Therefore, users can sense the earthquake by looking at its impact on the building elements and the movement of objects in the virtual environment, including the reception of auditory feedback during the event. Falling objects as well as debris formations can be 'easily' represented in the virtual environment by combining particle systems and the physical engine of the selected game engine [18].

Different solutions can be identified to include the physical shaking dimension in the gaming experience such as (1) shake tables, (2) seat or couch shaking systems, and (3) game controller vibration motors [46,47]. Shake tables represent the most advanced approach as they allow users to sense the shaking experience that they would experience in a real earthquake. However, despite the higher level of realism, such a solution has location constraints as it can only be placed in a laboratory and comes with associated costs. Another limitation is the possibility of injury by falling due to loss of balance during the gaming experience. Game controller vibration is the easiest option to setup without any location constraints. However, it provides a low level of realism as the game controller shaking can only provide a basic sense of vibration to users. Seat shaking represents a good compromise between shake tables and game controller vibrations. Seats and couches can be easily transported and can generate a relatively strong shaking experience for a seated user.

### 3.2 Building Selection and Representation

The game location represents the second key element to define for the VR SG prototypes. It is possible to select either existing buildings or new buildings [9]. The first approach provides users with a virtual environment they can be familiar with if users are building occupants of the existing building. This approach allows the investigation of whether building occupants are familiar with the evacuation plans and the evacuation routes, and the training of building occupants to enhance their evacuation preparedness in case of an earthquake. The same virtual environment can be used to investigate the behaviour of users who are not familiar with a given building to assess whether or not they follow the evacuation procedure as intended in the evacuation plans. However, this solution requires enough data to develop a virtual environment having features (i.e. indoor geometry, materials and pieces of furniture) fairly similar to the existing built environment [18,23].

In contrast, the second approach (the use of a hypothetical building) simplifies the process of creating a virtual environment since it does not require the collection of any real-world data [9]. However, the use of a hypothetical building does not provide the possibility of investigating the impact of occupants' familiarity with the building layout, which can be a key behavioural factor during building evacuations [48]. Different solutions can be used to make users familiar with hypothetical virtual environments, such as letting them navigate in it before the emergency evacuation is run within the VR SG framework. Another limitation of hypothetical buildings is that they do not have any evacuation plans, which can be a fundamental piece of information for design purposes (see Section 3.5). However, evacuation plans from existing buildings similar to the hypothetical one, or even evacuation recommendations provided by emergency regulatory agencies, can be used.

Depending on the selected building (i.e. hypothetical vs existing), different strategies can be used to represent the built environment and its geometry within the game structure. For instance, in the case of existing buildings, a comprehensive representation of the geometry of a building can be created by importing an existing BIM model into the game engine as discussed by Rüppel and Schatz [18]. This can provide highly detailed information (e.g. building element material, pieces of furniture, lighting conditions and evacuation systems), by integrating data identified through inspection. VR SG developers do, however, need to balance the amount of building information required for the gaming experience so as not to obtain portions of the building that are not visible during the game experience. In the case of hypothetical buildings, the representation can be inspired by existing BIM models of actual or new buildings. In either case, BIM models do not need to be extremely accurate representations of the building itself, but sufficiently realistic models that can provide users an appropriate sensory experience.

The use of BIM models to develop the virtual environment for VR SGs is a novel challenge in the field of construction informatics [49]. Regardless of the broad consensus amongst researchers that integrating BIM and gaming is worthwhile, several challenges and technical limitations have been reported by several authors [50–52]. One of the main technical limitations is that BIM models may contain large amounts of data and they tend to lose some parameter information, such as the material colour or texture, when converted and imported in a game engine. Different pipelines have been tested and proposed in the literature depending on the (1) data source, (2) 3D modelling software, (3) BIM software, (4) game engine, and (5) end-user platform. A review of those possible options is available in [52].

A final challenge is the representation of the environment surrounding the selected building as well as the external representation of the building itself. There may be external damage and hazards that users might need to be aware of before reaching designated safe places (e.g. assembly areas). In addition, there are the issues of how much of the external environment needs to be represented and at what level of detail.

### 3.3 Damage Representation
The representation of the earthquake damage to the building is another challenge that needs to be addressed when developing VR SG prototypes. An earthquake can impact the structural elements (i.e. components transferring the loads of the building to the ground), non-structural elements (i.e. components and systems that are permanently attached to structural elements but do not transfer loads) of a building, as well as the objects in the building, such as pieces of furniture. Building damage can be represented quantitatively or qualitatively [21].

Quantitative strategies consist of modelling all the structural elements of a building (i.e. geometry and materials) to investigate the impact of a hypothetical or recorded earthquake on these elements. Then, the damage to the non-structural elements and the systems can be predicted by using fragility functions, which are relationships between the probability of failure of a non-structural element and the damage to the structural elements. This approach has already been used to predict the impact of an earthquake on building evacuation using different evacuation simulation approaches [53,54]. The main limitation of the quantitative approach is that it requires advanced calculations as well as many pieces of information such as the structural and non-structural building components, earthquake location and magnitude, ground motion conditions, etc. The number of assumptions that are necessary for those calculations can be large enough to reduce significantly the reliability of the damage representation [21].

While such calculations are fundamental for simulation studies, they are not essential for the development of a SG since building occupants are not aware of how the building would behave in the case of an earthquake. A realistic representation of the building damage in SGs can be created using a qualitative strategy. Such a strategy is based on mimicking the damage in the virtual environment by using data from existing datasets including videos and photos of building damage [55]. This can be considered a valid approach as the realism of building damage should be in line with the expectations of users. The qualitative approach leaves the SG developers free to locate the damage in 'strategic positions' of the building for specific training purposes and to investigate specific evacuation behaviours. By using the qualitative strategy, it is still possible to assume and represent realistic levels of damage for pieces of furniture in the building. As a result, this approach helps to establish a credible and reliable experience for users rather than modelling earthquake damage on a building accurately which can be resource-intensive.

### 3.4 NPC Behaviour and Participant-NPC Interactions

A VR SG prototype can enable investigation of the ways in which participant behaviour is influenced by other evacuees. This can be done by using multi-participant games, or introducing NPCs, or a combination of the two. The first solution allows several users to inhabit and interact in the same virtual environment. However, the number of users can be strongly limited by the VR hardware, for instance the number of available HMDs and the graphics card performance. The second solution does not have this limitation as many NPCs can be modelled as agents. Representing evacuation NPCs in a virtual environment is a further challenge for the development of a VR SG prototype. Depending on the outcome aimed for the decision also needs to be made around whether NPCs behave realistically (following existing data on human behaviour in earthquake), or whether they behave according to best practice (if used as part of a training intervention to improve behaviour). Qualitative data and quantitative data (e.g. walking speeds and evacuation flows) on earthquake evacuation behaviour can be identified in the existing literature. These data have been collected using different techniques, namely video analysis [55–57]; interviews and questionnaires [58–60]; direct observations during an earthquake [61]; and evacuation drills [62,63]. Despite the lack of a comprehensive literature review on human behaviour in earthquake evacuations, several key behavioural statements and qualitative data can be identified and used to program NPC behaviour during an earthquake. A summary of some general behavioural assumptions to take into account to define NPC behaviours is presented proposed in [4,57].

In reaction to the building damage, the behaviour of NPCs should be realistic enough, but it does not need to represent the actual behaviour that building occupants would have in a specific scenario. In fact, different NPCs responses can be included in the VR SG prototypes to investigate how users are affected by social interactions with NPCs. Some NPCs might follow either behaviour recommended by existing guidelines or perform unsafe actions, such as to start evacuating during the earthquake. The choice between these options depends on the purpose of the VR simulation in the prototype: behavioural study; or training best evacuation practices.

### 3.5 Behavioural and Learning Outcomes

Defining the outcomes is a crucial step in developing VR SGs for earthquake evacuations, since a SG should provide learning outcomes to users as well as behavioural outcomes for the SG developers.

The learning goal of post-earthquake evacuation is to educate users how to respond during the main shock and how to evacuate safely from a building damaged by an earthquake. This can be done by identifying the appropriate and inappropriate behaviours that evacuees can have in earthquake evacuations and correcting them. For a list of those behaviours, designers can refer to international or national guidelines on the best practices or the evacuation plan of the buildings represented in the VR SGs. Another challenge to consider is how to weight these behaviours in terms of importance to life safety outcomes.

Once these behaviours have been identified a fundamental question is: how can feedback be provided to users? Several solutions have been used to date in existing SGs for evacuation training. For instance, some SGs use several non-verbal vocal sounds of human distress, sounds of breaking bones, blood squirts, simulation of temporary blindness, tinnitus or dizziness and a life bar (i.e. a bar showing the health of the virtual participant in the game) to highlight the severity of the consequence of the unsafe actions taken by SG users, coupled with a cause of death message (i.e. an in-game text message indicates why the users lost their lives in the game) and a behavioural recommendation [19,20,64]. Less extreme approaches include the use of 'knowledge points' for correct decisions and the loss of points or time penalties for incorrect decisions [20,65]. In many of these 'death' scenarios, users are taken back to the point before their incorrect decision took place and allowed to replay from that point. The more extreme methods discussed above have the advantage of adding realism, potentially eliciting a motivational response some degrees closer to what they would experience during an actual earthquake; however, there are ethical considerations in countries where a portion of the population has been affected by major earthquakes, which may preclude their use in favour of lower fidelity options.

### 4. CASE STUDY

In this paper, we present the Behavioural Prototype (BP) and Training Prototype (TP) developed to investigate human behaviour in earthquake evacuations and to train users the best practice on how to cope with an earthquake according to the New Zealand Civil Defence guidelines [2]. The design choices and steps followed for a VR SG prototype development are described in the following subsections.

This section represents a possible instance of VR SGs generated from the general concepts and solutions described in Section 3. The proposed prototype is the result of interaction between researchers with diverse backgrounds (i.e. Civil Engineering, Computer Science and Social Science) and advisors from several public organizations such as New Zealand Civil Defence and the Auckland District Health Board.

### 4.1 Virtual Environment

The game location for both prototypes is a section of the Auckland City Hospital (ACH) which is New Zealand's largest public hospital and clinical research facility. This location was selected as it is a key facility in Auckland that gives access to several categories of building occupants such as staff members (i.e. medical staff, administrative staff) and visitors. Therefore, the BP has the potential to investigate how these different occupant groups respond to earthquakes. The TP enables the assessment of the effectiveness of the SG in comparison with traditional approaches such as health and safety inductions, leaflets, and seminars. Moreover, the importance of selecting a hospital as a case study is related to the need for alternative solutions to train building occupants and to investigate human behaviour as traditional drills cannot always be carried out in such buildings, given the presence of vulnerable populations [6]. To the best of our knowledge, only two serious games have been developed for hospital scenarios, both of which refer to fire evacuations [20,66]. As such, the proposed prototypes are the first implementation of a SG for post-earthquake evacuation training implemented in a hospital environment.

The public and administrative areas located on the fifth floor of the ACH were chosen to develop the BP and TP as both visitors and staff members are familiar with at least some areas of that floor. The 3D model of this section of the building was developed by using BIM software tools. The main input was pre-existing BIM models, 2D DWG files, site visits and photographs taken of the building by the researchers. First, a new BIM model of the building was created using Autodesk Revit and imported into Unity 5. The second step of the pipeline consisted of exporting the BIM model from Revit and convert it to the FBX format, which is directly readable by Unity. The final adjustments of the 3D model of the building were made in Unity. The final step of the pipeline consists of making the door, lift and escalator dynamics using C# scripts to allow participants to interact with those objects during the game. In this step, final items were added to the virtual environments such as stationary and specific pieces of hospital furniture not available in the Revit accessories. Figure 2 shows the pipeline described in this paragraph. Figure 3 shows a top view of the modelled area in Unity while Figure 4 juxtaposes model views and photographs.

A final challenge was the optimization of the virtual environment for the gaming purpose. Ideally, a VR application should run at a frame rate equal to or greater than the HMD refresh rate. As such, it was necessary to optimize the rendering settings of the virtual environments to get as close as possible to the ideal target of 50 frames per second [38]. To achieve such a goal, Lightmapping and Occlusion Culling algorithms were used. Lightmapping is a form of surface caching in which the lighting characteristics of surfaces in a virtual scene are pre-calculated and stored in texture maps instead of being calculated in real time. The Occlusion Culling algorithm disables rendering of objects when they are not currently seen by the camera as they are entirely obscured by nearer objects.

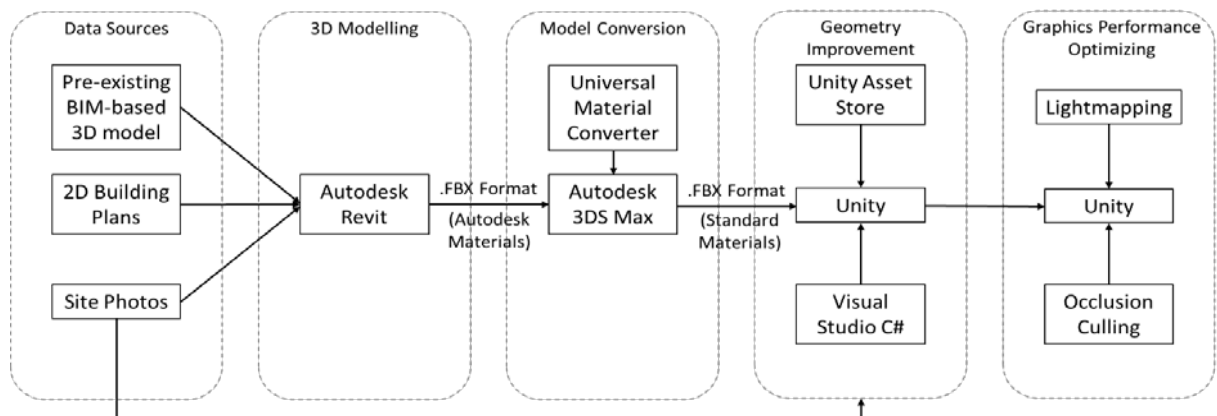

Figure 2 – Development pipeline implemented to create the virtual environment for the BP and TP

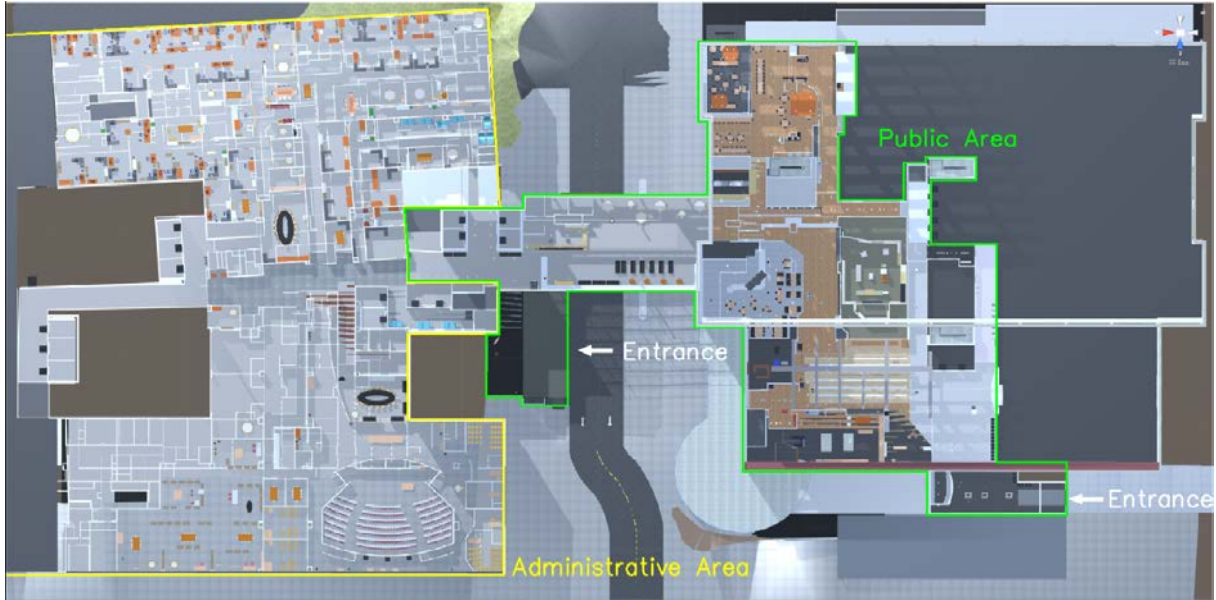

Figure 3 – Top view of the portion of the fifth floor of ACH modelled in Unity

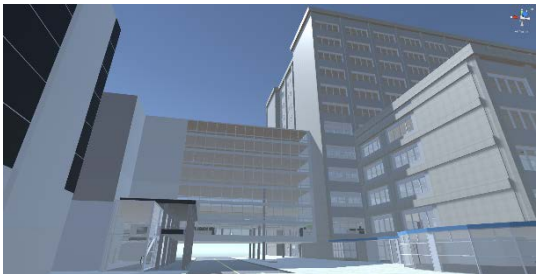
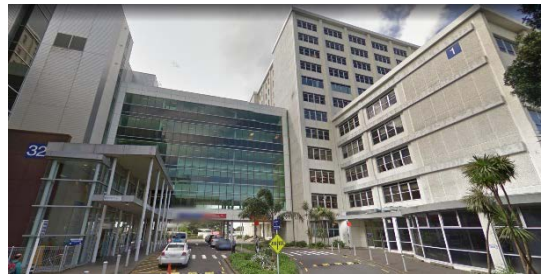
(a) (b)
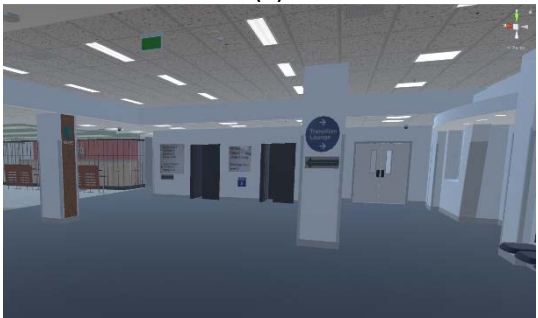
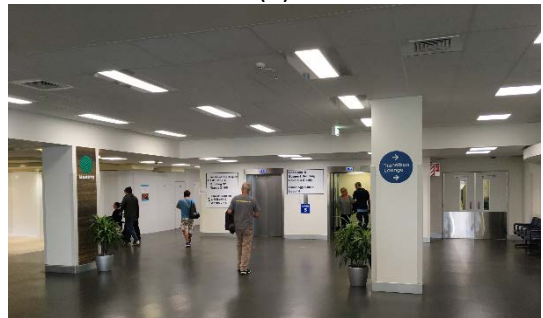
(c) (d)

Figure 4 – Comparison between virtual model and photographs of ACH.

### 4.2 Earthquake Simulation and Building Damage
The earthquake and the related building damage were simulated in both prototypes using a qualitative approach, as described in Section 3.3.

The earthquake was simulated by shaking the floor of the virtual room in which users were located. The setting of the earthquake code was selected to generate earthquake damage equivalent to the VII-VIII intensity of the Modified Mercalli Scale. To simulate the earthquake, the floor was shaken as a single entity along the same direction at any time while the motions of the objects in the scene are then computed and updated at each frame by the physics engine. The shaking propagation from the floor to the objects in the room is illustrated in Figure 5.

For each frame, the Unity physics engine applies a virtual force to the floor according to the direction and intensity features generated by the 'Earthquake Generator' package. The impact of this virtual force on the objects on the floor is then calculated by the physics engine depending on the mass and the static or kinematic friction coefficients of those objects and on the sliding condition of those objects. Considering the example in Figure 5, $F_{table}^{floor}$ is the force that the floor acts on the table while $F_{floor}^{table}$ is the force received by the floor from the table. Those forces are calculated using the static friction coefficient if the table is not sliding on the floor and kinematic friction coefficient otherwise. The same interaction occurs between the table and the box though $F_{table}^{box}$ and $F_{box}^{table}$. The physics engine also generates a virtual gravity force that allows objects to fall or to overturn when their equilibrium conditions become unstable.

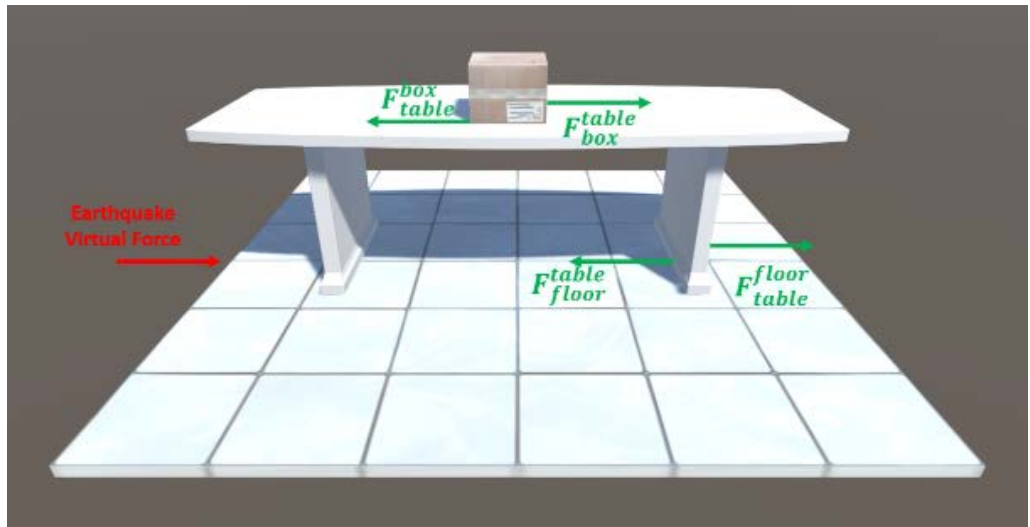

Figure 5 – Earthquake force propagation in the virtual environment

The earthquake is only simulated in the areas where the participant is located as the dynamic generation of the earthquake impact on objects described in the previous paragraph is computationally demanding. The earthquake impact in the remaining parts of the building is programmatically generated using a C# script that relocates virtual objects using the final locations list generated by a pre-simulated scenario. In other words, the impact of the earthquake on the objects in the remaining part of the building was generated once for all scenarios and saved in a list of the final positions of those objects. During the game, the C# script accesses this list and relocates those objects using co-routines to avoid performance overhead during the simulation.

The building damage was programmatically generated using a qualitative approach though a C# script that replaces the undamaged walls and ceiling with damaged walls and ceiling. The ACH has drywall partition walls while the false ceiling is a plaster ceiling tile system. Damage to boards and tiles were added to the building during the earthquake stage. The comparison between the damaged and undamaged scenarios is illustrated in Figure 6.

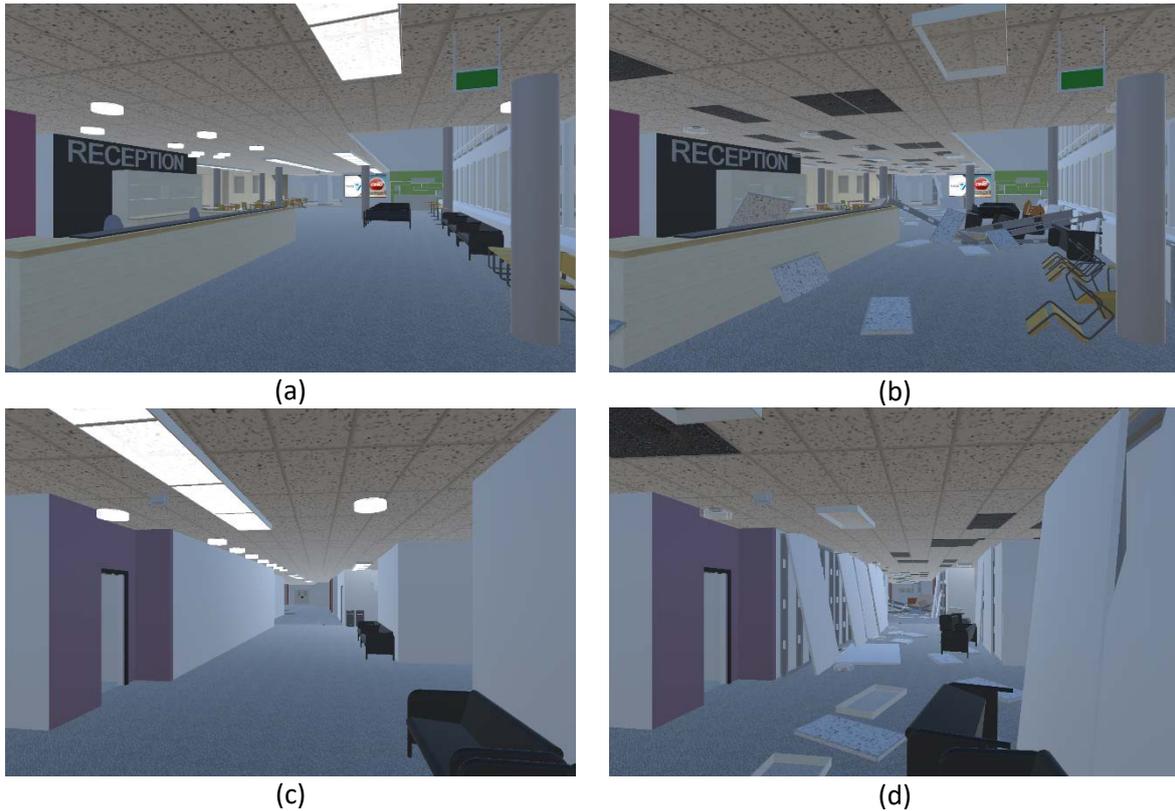

Figure 6 – Comparison between the undamaged environment (a and c) and damaged environment (b and d)

Finally, the physical shaking dimension for the VR experience was implemented using a chair shaking system, which consisted of a vibrating platform on which users were seated during the VR experience. A 'ButtKicker Gamer' was installed in the platform [47]. This haptic system was activated by the earthquake noise generated during the game and it vibrated following the low frequencies of this noise. A photograph of a tester using the haptic system is illustrated in Figure 7.

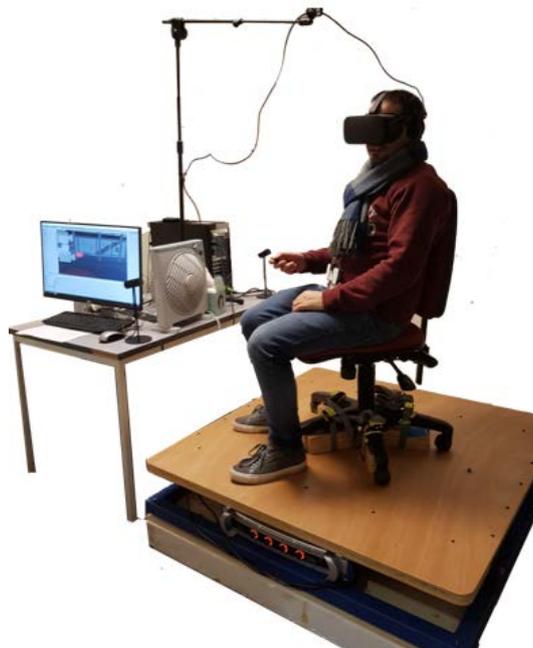

Figure 7 – A tester seated on the vibrating platform

### 4.3 Non-player Characters

Non-player Characters (NPCs) are the characters not controlled by the game participant. Their behaviours are predetermined by scripts. The NPCs are used to make up the population in the ACH virtual environment. There are two types of NPCs in the game - non-interactive and interactive NPCs.

The non-interactive NPCs are located all the way along participants' navigation areas in the virtual environment. There are different pre-assigned roles to make the virtual environment more realistic, including female and male, young and old, staff and visitors, patients and medical personnel. The NPCs are walking, talking, standing, sitting, eating or drinking following a typical daily routine in the ACH in the pre-earthquake part of the SG. There are several other interactive NPCs used to guide or accompany participants, which may influence participants' behaviour during and after the earthquake. The interactive NPCs are used to investigate social behaviours which are typical during building evacuations [34,67,68].

A pipeline was developed to create the NPCs and animate them using the available software packages illustrated in Figure 8. The NPCs were first created in Adobe Fuse CC, which is a software for creating 3D human models and NPCs. Then the models were uploaded to Adobe Mixamo for skeleton rigging and animating. Mixamo is a 3D character cloud-based animation software that contains several animations in its library, allowing rapid animation development. After rigging and animating, the models were exported in FBX format and then imported to Unity. Where the desired animations (such as NPC under debris) were unavailable in the Mixamo llibrary, they were drawn from the Unity Asset Store.

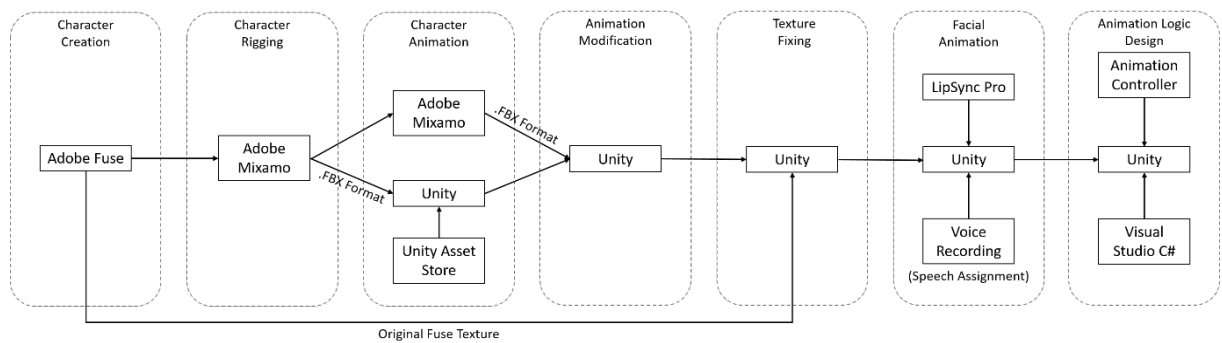

Figure 8 – Development pipeline implemented to develop and animate NPCs.

To make the NPCs more realistic, a plugin called LipSync Pro in Unity Asset Store was used to make facial animations. The NPC lips moved when they were speaking and blinked like real human beings. Furthermore, NPCs were assigned texts where necessary. The non-interactive NPCs speak during the full game without interacting with the participants while the interactive NPCs speak whenever events are trigged by the participants (e.g. entering in a room, taking an action, etc.). The final step was to add C# scripts and animation controllers to control the NPCs' animations, including non-interactive and interactive animations. In this step, the NPCs were assigned in each specific scenario and the animations were triggered under certain conditions so that they became active in the game.

### 4.4 VR Navigation Solutions

The navigation solution adopted in the BP is a combination of open world solution and a first-person controller. This solution allows participants to move freely through a virtual world with a given level of freedom regarding how and when to approach objectives to follow the designed story line (see Section 4.5). One limitation of an open world solution is that it requires many post development checks to make sure that participants do not get stuck in any points of the virtual environment. The sickness related to experiences of self-motion in the absence of actual physical displacement is another one of the challenges to face in VR open worlds. One of the most popular solutions to reduce this issue is teleportation [38]. This solution allows participants to teleport through the experience from one location to another, rather than walking. This solution reduces to milliseconds the experiences of self-motion in the virtual environments but it breaks the realism of the experience, which is a fundamental feature for the BP. We prefer instead the navigation solution in which participants can navigate toward the direction they are looking at by clicking a single button. Using this solution, participants are expected to rotate with their full body accordingly towards the direction they want to go in the virtual environment. This solution reduces the sickness as the rotations in the virtual and physical environments are the same. To avoid the risk of participants falling while rotating, they were asked to sit on a rotatable chair giving them the possibility to rotate through a full 360º as illustrated in Figure 7.

The navigation solution adopted in the TP is a combination of wait points and a first-person controller. This solution allows participants to move through predefined scenarios following a planned order where they need to choose an action to proceed to the next scenario. In this case, the navigation between wait points is driven by a C# navigation script taking participants through planned trajectories. The selection of actions is done by choosing among several action panels as illustrated in Figure 9. Participants can select action panels by looking at them and clicking a single button. The main limitation of a wait point solution is that it reduces the possibility of participants to navigate in the environment and thus the ecological validity of the behavioural data. However, this did not represent a concern for the development of the TP as it is mainly designed to generate learning outcomes rather than being a fully interactive environment.

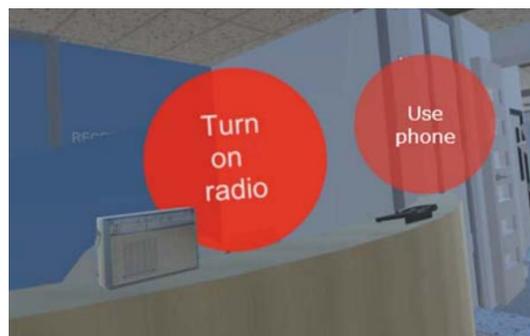

Figure 9 – An example of two action panels visualized by participants in the TP

**4.5 Prototype Outcomes and Story Lines**
The main difference between the BP and TP is the design and selection of behavioural and learning outcomes.

The BP was mainly designed to generate behavioural outcomes. As such, participants were not provided with any feedback or instructions regarding the best practice as that could affect their behaviours. The BP was designed to answer specific research questions by observing the participants' interaction with the virtual environment during and after the simulated earthquake. We defined a list of questions that allowed for the quantification of the type, the priority and the duration of the actions taken during distinct phases of the earthquake emergency: earthquake phase; pre-evacuation phase; indoor evacuation phase; and outdoor evacuation phase. The key behavioural questions leading our BP design are summarized in Table 1. To answer those questions, participants are exposed to a virtual environment that allowed them to act according to the best evacuation practice if possible or to take unsafe actions. Many of those behaviours are recommended by the New Zealand Civil Defence [2] and the ACH evacuation plan [69] as indicated in Table 1.

Table 1 – Behavioural questions leading the BP design

| Earthquake Phase | Recommended Behaviour |
|---|---|
| Do participants drop, cover and hold (DCH) as the first action? | Yes |
| Do they have alternative behaviours as the first action? If yes which ones? | NA |
| How long did s/he take to decide to DCH from the start of the earthquake? | NA |
| **Pre-evacuation Phase** | |
| How long do they take to get out from under the table (if under)? | NA |
| Do they check for damage? | Yes |
| Do they unplug broken electronic device? | Yes |
| Do they use the phone to call someone? | No |
| Do they use the phone to text someone or to browse for information? | Yes |
| Do they assist other people? | Yes |
| Do they use the radio to collect information? | Yes |
| Do they take or use a first aid kit? | Yes |
| Do they use computers to browse for information? | Yes |
| Do they collect personal belongings? | Yes |
| Do they wait for instruction before starting evacuating? | Yes* |
| How long do they wait before exiting the room after the earthquake? | NA |
| **Indoor Evacuation Phase** | |
| Do they check the damage while evacuating? | Yes |
| Do they use the stairs or escalators? | Yes |
| Do they use lifts? | No |
| Do they check for injured people before going downstairs? | Yes |
| Do they check the damage of the stairs or escalator before using them? | Yes |
| **Outdoor Evacuation Phase** | |
| Do they stay close to the building? | No |
| Do they return inside the building? | No |
| Do they identify a safe earthquake assembly area? | Yes |

*for visitors

The TP was designed to generate training outcomes, i.e. enhance participants' knowledge on how to behave during and after an earthquake. Starting from the guidelines by New Zealand Civil Defence [2] and the ACH evacuation plan [69], we identify a list of recommended behaviours that needed to be taught through the TP as illustrated in Table 2. During the TP, participants were guided through several scenarios in which they needed to choose one or more of the recommended behaviours in Table 2 or alternative behaviours which are not in line with best practice.

The main goal driving the definition of the story lines for the BP and TP were the behavioural and training outcomes described in Table 1 and 2. However, the definition of these story lines needed to take into account all the constraints and possibilities related to the other components of the SG, such as the geometry of the virtual environment (see Section 4.1), the earthquake and damage simulations (see Section 4.2), the presence of NPCs (see Section 4.3) and the navigation solutions (see Section 4.4).

It was necessary to create a scenario in which participants could drop, cover and hold in a room during the earthquake phase and to perform all the potential actions described in the pre-evacuation phase included in Table 1. Finally, the evacuation path in the building and out of the buildings needed to have game components aimed at investigation of the actions in the indoor and outdoor evacuation phases stated in Table 1.

Table 2 – Recommended behaviours included in the TP

| Earthquake Phase |
|---|
| Drop, cover and hold or take cover under table |
| Pay attention to falling, objects or sharp objects such as broken glass |
| **Pre-evacuation Phase and Indoor Evacuation Phase** |
| Wait for 30s after shaking to see if there's any more shaking due to an aftershock |
| Collect personal belongings |
| Take first aid-kit |
| Check and help people around |
| Search for available exit if usual or closest one is blocked |
| Put out fire with a fire extinguisher or report to fire service |
| Unplug broken electrical equipment |
| Listen to the radio to collect information |
| Use stairs to exit |
| **Outdoor Evacuation Phase** |
| Stay at an open space away from buildings, falling objects and glasses |

Considering all the needs and constraints for the BP, the following story line was used:
1. Participants start the game outside the ACH and they are asked to reach a meeting room in the hospital by navigating a small section of the public and administration area.
2. Once participants reach the room, they are invited to leave their belongings on a table by two NPCs (i.e. a doctor and another visitor). Participants leave their smartphone and their keys on the table.
3. As they leave their belongings, the earthquake starts. Participants can drop and cover under the table or take alternative actions, such as trying to escape from the room. During the earthquake, the two NPCs in the room take cover under the table until the end of the earthquake.
4. Once the earthquake has finished, the doctor NPC leaves the room to assess the situation and asks the participants to wait for instructions. At this stage of the game, participants can take multiple actions by:
    a) checking the damage in the room;
    b) unplugging a damaged printer;
    c) using the phone to call, to text or browse for information;
    d) assisting the NPC character that is still in the room;
    e) using a radio in the room to collect information;
    f) collecting and using a first aid kit;
    g) using a laptop in the room to collect information from the internet;
    h) collecting personal belongings;
    i) start evacuating.
5. Participants can start evacuating before or after receiving instructions from the doctor NPC. Once the indoor evacuation is started, they can:
    a) assess the damage while evacuating;
    b) choose between stairs, stopped escalator or lift;
    c) check the damage of the stairs or escalator before using them;
    d) check for injured people;
6. Participants can finally evacuate the building and they can finally look for a safe assembly point or return to the building.

The story line for the TP was designed to train participants according to the training outcomes defined in Table 2. The main constraint for this prototype was the need to simulate earthquake damage only for a limited area of the building. As such, we constrained the story line to have the participant in a defined room when the earthquake strikes, as for the BP. In this case, there was no need to adjust the playable area as the wait points solution was adopted and the navigation was run by code.

The story line for the TP consists of the following points:
1. Participants start the game outside the ACH and they are asked to reach a meeting room in the hospital by following a staff member.
2. Once participants have reached the meeting room, they are welcomed by a doctor NPC, who invites them to leave their belongings on the table.
3. As they leave their belongings, the earthquake starts and participants can drop, cover and hold under a table or beside unsafe objects, or take other unsafe actions available in the scenario.
4. When the earthquake ends, the doctor in the room leaves to check the situation while participants can take several safe and unsafe actions available in the scenario. The recommended actions included in this part of the TP are:
    a) Stay under cover and wait for 30 seconds;
    b) Collect their belongings;
    c) Take the first aid kit in the room.
5. Finally, participants have the option to get out of the room to start evacuating. While evacuating participants find several scenarios in which they can take the following recommended actions:
    a) Assist a nurse to help an injured NPC;
    b) Help an NPC stuck under the debris;
    c) Search for an available exit if the usual or closest one is blocked;
    d) Unplug broken electrical equipment;
    e) Extinguish a small fire or report it;
    f) Listen to the radio to collect information;
    g) Find a safe way out of the building by stairs;
6. Participants reach the exit of the building and then they can choose a safe assembly area.
7. The experience ends in a virtual environment where participants receive a report showing all the recommended actions that have been taken. A video and audio guide take participants through all the choices they made in the game, explaining the rationale behind every recommended action.

### 4.6 Data Collection and Participants
The two prototypes were tested with staff members and visitors at the Auckland City Hospital from July 17 to August 4, 2017. The participants were contacted by emails using email lists and using leaflets and posters spread in the hospital. 179 participants decided to take part to the experiment. However only 170 participants completed the experiment 54 were staff members and 116 were visitors. The remaining 9 participants stopped the VR experience as they felt motion sickness. Each participant was randomly assigned to one of the two prototypes: 83 participated tested the BP while 87 participants tested the TP.

The participants followed two different experimental protocols depending on the prototype they were assigned to. BP participants were asked to read the information form and sign the consent form. Once instructed about the VR equipment and trained on how to use the VR navigation system, they started the experiment. As the experiment was concluded participants were asked to fill out a questionnaire asking for their demographics and to assess their VR experience.

TP participants were asked to read the information form and sign the consent form. They were asked to fill out a first questionnaire asking for their demographic information. Then, they were interviewed to assess their earthquake preparedness. Once instructed about the VR equipment and trained on how to use the VR navigation system, they started the experiment. Then, they were asked to fill out a second questionnaire to assess their VR experience.

The statistics of the BP and TP participants are summarized in Figure 10. It is possible to observe that the demographics of the two groups are similar. Moreover, the sample of this study shows a high variance of the participants' age which is not a common feature in many previous studies as most of the participants of those previous studies were university students [70].

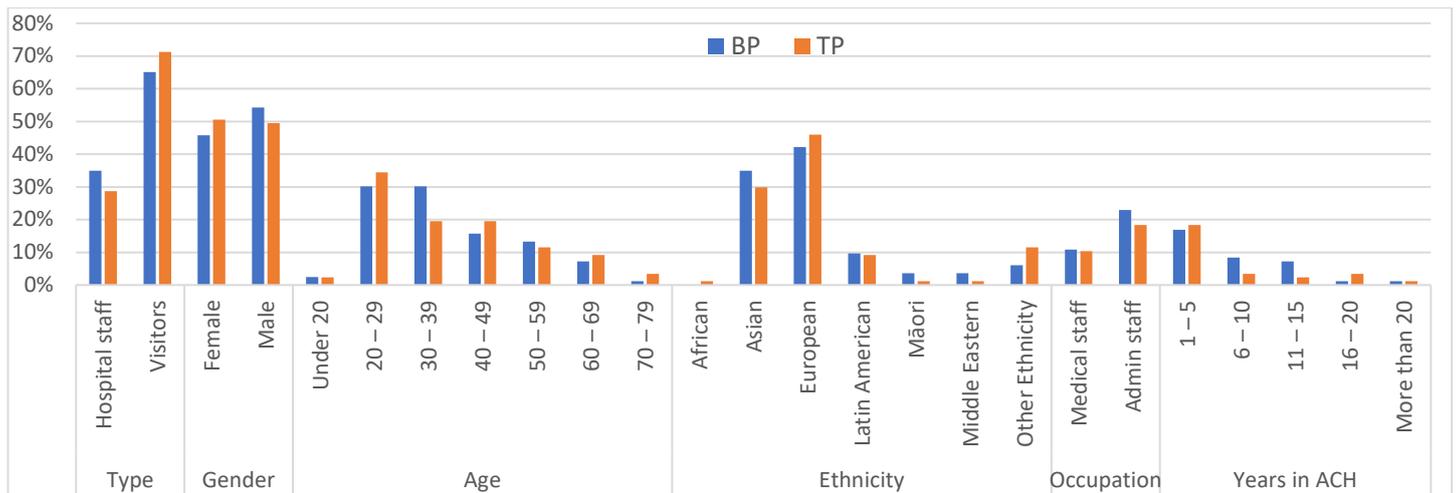

Figure 10 – Demographics of the BP and TP participants
(Occupation and Years in ACH refers to staff members)

### 4.7 Prototype Assessment and Validation

The components of the BP and TP were assessed by the participants using a Likert scale questionnaire (-3 very low; +3 very high). Participants were asked to assess the following components:

1. Realism of the virtual environment;
2. Perceived realism of earthquake simulation and damage;
3. Realism of the NPCs;
4. Game navigability and interactivity.

The average score provided for the two prototypes are illustrated in Table 3. The results indicate that all the component receive a positive score. Those results are fundamental to engage participants and let them 'believe' on the credibility of the threat and related outcomes. The component having the lowest average score is the realism of the NPCs. This indicates that more efforts are necessary in future NPC design. The Wilcoxon Signed Rank Test was used to compare the score of the two prototypes for each component. The results in Table 3 indicate that there is not a significant difference.

### 5. DISCUSSION AND CONCLUSIONS

This paper addresses the prototyping issues and challenges for SGs related to earthquake emergencies. Although other works discussing prototyping challenges for emergency SGs are available in the literature, those focus mainly on fire evacuations or other firefighting procedures. As such, we identify the key components necessary to develop a VR SG for earthquake emergencies (earthquake features, building selection and representation, damage representation, NPC Behaviour and Participant-NPC Interactions and behavioural and learning outcomes) and possible developing solutions. In this paper, we also introduced the case study of the ACH showing the practical steps we followed to develop two VR SG prototypes to improve earthquake preparedness in a public building in New Zealand. Through this case study we proposed and discussed possible development pipelines and design solutions to answer our original research questions. The perception of the design components is investigated by collecting data from 170 participants working at or visiting the ACH.

Table 3 – Participant assessment of the BP and TP components. (Data from Likert scale questions: -3 very low; +3 very high)

| Component | Prototype | Mean | Std. Dev. | P-Value* |
|---|---|---|---|---|
| Realism of the virtual environment | BP | 1.95 | 0.99 | 0.607 |
| | TP | 1.84 | 1.16 | |
| Realism of earthquake simulation and damage | BP | 1.82 | 0.97 | 0.690 |
| | TP | 1.80 | 1.17 | |
| Realism of the NPCs | BP | 1.06 | 1.49 | 0.212 |
| | TP | 1.40 | 1.37 | |
| Game navigability and interactivity | BP | 1.76 | 1.36 | 0.167 |
| | TP | 2.07 | 1.16 | |

*Wilcoxon Signed Rank Test

In the ACH case study, we presented a possible solution based on the use of the Unity physics engine. The proposed solution identifies the need for a compromise between the computational cost of a dynamic simulation of an earthquake and the level of realism. Another challenge of earthquake SGs is the simulation of the building damage. Unlike a fire emergency, the building damage generated by an earthquake has an impact on the full building at the same time. In other words, it is necessary to apply the damage to the full virtual environment at the same time during the shaking stage. In the case study, we show how this can be done by combining a qualitative approach and co-routine scripts in C#. The realism of the proposed solutions was assessed by asking participants to assess the realism of the virtual environment and earthquake simulation and damage. Results in Table 3 indicates that the main perception of those two components is relatively high for both prototypes.

A significant design difference of earthquake SGs from fire SGs is that the list of recommended actions is much longer and complex than the one for fire emergencies. In fact, while the main goal in fire emergencies is to evacuate immediately according to the paradigm 'the sooner the better', in the case of earthquake emergencies, several pre-evacuation tasks are suggested to carry out a safe evacuation (see, for example, Table 1 and 2). This means that participants capabilities to handle an earthquake emergency cannot be assessed using the total evacuation time, which is instead one of the main indicators for fire SGs. Moreover, the complexity of the earthquake evacuation procedure can have a significant impact on the design of the story line. Our case study clearly shows how the story lines for earthquake SGs can be much longer and more articulate than the ones implemented in classic fire SGs.

NPCs are key components for many emergency SGs as they allow the investigation of social interactions during emergencies. The importance of NPCs has been highlighted by previous studies [20,23,66]. In this work we highlight that NPCs are still key components for earthquake emergencies as social interactions are fundamental to train participants and to investigate their behaviours. The perception of NPCs in the two prototypes were assessed by the participants. The results in Table 3 indicates that, although there is a positive level of perceived realism, more efforts are necessary to increase the realism. In fact, this component has the lowest score among those in Table 3. This is probably due to the fact that NPCs were trigged in only few specific events and a comprehensive interaction between NPCs and participants was not developed in the proposed prototypes. As such, the refinement of the NPCs represents a future challenge for the development of a second generation of SGs.

Another practical challenge that a VR SG designer needs to overcome is the reduction of the motion sickness related to experiences of self-motion in the absence of actual physical displacement. Different VR navigation solutions have been proposed to mitigate this issue. Though the ACH case study, we identified two navigation solutions depending on the goal of the SG. To enhance the realism and ecological validity of the BP we propose an open world solution. In the BP, participants need to rotate with is full body accordingly towards the direction they want to go to in the virtual environment and they can navigate toward the direction they are looking at by clicking a single button. By contrast, the game realism is not a main objective for the TP while it is necessary for participants to go through predefined choice scenarios in which they can to select recommended actions or avoid them with a more rigid story line. As such, we identified the wait point solution as the most appropriate navigation solution. In this case a single button can be clicked to select the action panels in each choice scenario. The only invariant between the BP and TP is that participants are asked to use a single button instead of using a classic game controller. This design choice was made considering that SG participants in a public building are not all gamers familiar with standard controllers. Therefore, we simplify the game controller as much as possible, reducing it to a single button. These navigation solutions were also assessed by the BP and TP participants showing that both solutions have a high level of usability and there is no statistical difference in the scores of the two systems. Moreover, only 5% participants felt motion sickness. This result confirms the high quality of the navigation systems. As such, new VR SGs could benefits from the implementation of those systems.

**ACKNOWLEDGMENTS**

This research has been funded by the MBIE-Natural Hazards Research Platform (New Zealand), Grant Number: C05X0907. The authors thank Andrew George, Bunpor Taing, Hakshay Kumar and Matthew Richards undergraduate students in Civil Engineering at The University of Auckland, for the development of the ACH BIM model and Bruce Rooke for the software development.